\documentclass{bmvc2k}

\usepackage{multirow}

\title{ Deep Spatio-temporal Manifold Network for Action Recognition}

\addauthor{Ce Li}{celi@cumtb.edu.cn}{1}
\addauthor{Chen Chen}{chenchen870713@gmail.com}{2}
\addauthor{Baochang Zhang}{bczhang@139.com}{3}
\addauthor{Qixiang Ye}{qxye@ucas.ac.cn}{4}
\addauthor{Jungong Han}{jungonghan77@gmail.com}{5}
\addauthor{Rongrong Ji}{rrji@xmu.edu.cn}{6}

\addinstitution{
	Department of Computer Science\\
	China University of Mining \& Technology, Beijing, China
}
\addinstitution{
	Center for Research in Computer Vision (CRCV)\\
	University of Central Florida, Orlando, FL, USA
}
\addinstitution{
	School of Automation Science and electrical engineering\\
	Beihang University, Beijing, China
}
\addinstitution{
	University of Chinese Academy of Sciences\\
	Beijing, China
}
\addinstitution{
	Nortumbria Univesity\\
	Newcastle, UK
}
\addinstitution{
	Xiamen Univesity\\
	Xiamen, China
}

\runninghead{}{Deep Spatio-temporal Manifold Network for Action Recognition}

\def\etal{\emph{et al}\bmvaOneDot}
\usepackage{times}
\usepackage{epsfig}
\usepackage{graphicx}
\usepackage{amsmath}
\usepackage{amssymb}

\newcounter{defcounter}
\usepackage{latexsym}
\usepackage{amssymb,amsmath,amsfonts}
\usepackage{algorithmic}
\usepackage{color}
\usepackage{url}

\usepackage[ruled]{algorithm2e}

\newcommand{\cut}[1]{}

\DeclareRobustCommand\onedot{\futurelet\@let@token\@onedot}
\def\@onedot{\ifx\@let@token.\else.\null\fi\xspace}
\def\etal{\emph{et al.}}

\begin{document}
	
\maketitle
	
\begin{abstract}
Visual data such as videos are often sampled from
complex manifold. We propose leveraging
the manifold structure to constrain the deep
action feature learning, thereby minimizing the
intra-class variations in the feature space and alleviating
the over-fitting problem. Considering that
manifold can be transferred, layer by layer, from
the data domain to the deep features, the manifold
priori is posed from the top layer into the back
propagation learning procedure of convolutional
neural network (CNN). The resulting algorithm --Spatio-Temporal Manifold Network-- is solved
with the efficient Alternating Direction Method of
Multipliers and Backward Propagation (ADMM-BP).
We theoretically show that STMN recasts the
problem as projection over the manifold via an embedding
method. The proposed approach is evaluated
on two benchmark datasets, showing significant
improvements to the baselines.
\end{abstract}
	
\section{Introduction}\label{Sec.1}

\begin{figure}
	\begin{center}
		\includegraphics[width=0.5\columnwidth]{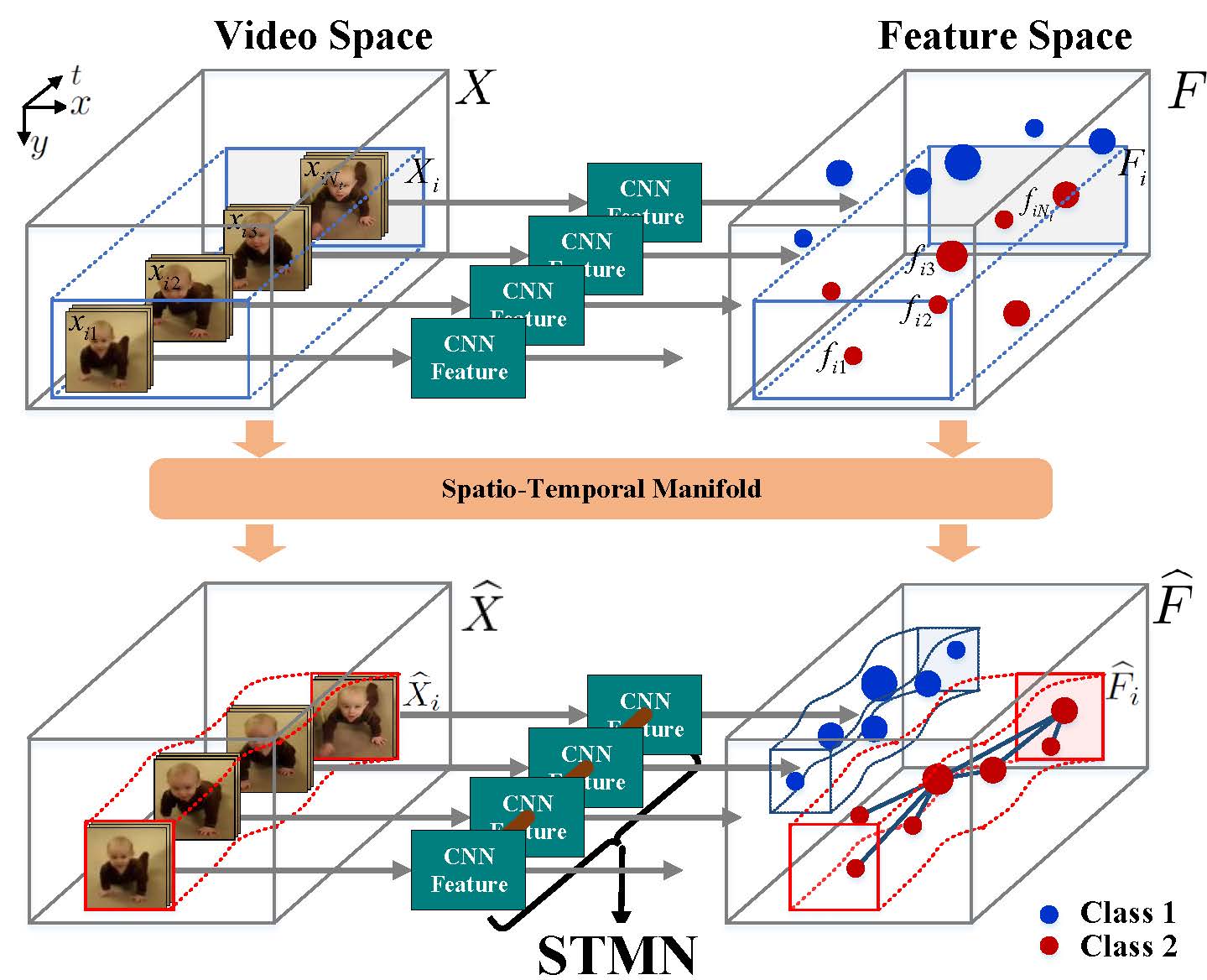}
	\end{center}
	\vspace{-1em}
	\caption{The STMN framework.
We model the intra-class action space as a spatio-temporal manifold, which is used as a regularization term in the loss function. Consequently, the manifold structure of intra-class actions remains in the resulting STMN approach. Two classes (blue/red) of samples in the CNN feature space are randomly distributed (upper), differently in STMN the manifold structure regularizes the samples in a compact space (bottom).}
	\label{fig.framework}
\end{figure}


Deep learning approaches, e.g. 3D CNN~\cite{Ji2013PAMI}, two-stream CNN~\cite{Simoyan2014NIPS}, C3D~\cite{Tran2015CVPR}, TDD~\cite{Wang2015CVPR} and TSN~\cite{Wang2016ECCV}, have shown state-of-the-art performances in video action recognition. Nevertheless, deep learning is limited when dealing with the real-world action recognition problem.
One major reason is that deep CNNs tend to suffer from the over-fitting problem~\cite{Szegedy2015CVPR}, when the labeled action ground truth is inadequate due to its expensive labor burden; and there exist significant intra-class variations in training data including the change of human poses, viewpoints and backgrounds.
Unfortunately, most of the deep learning methods aim at distinguishing the inter-class variability, but often ignore the intra-class distribution ~\cite{Lu2015CVPR}.

To alleviate the over-fitting problem, regularization techniques~\cite{Lu2015CVPR} and prior knowledge (e.g. 2D topological structure of input data~\cite{Le2011CVPR}) are used for image classification task.
Nonlinear structures, e.g. Riemanian manifold, have been successfully incorporated as constraints to balance the learned model~\cite{Zhang2015CVPR,Wen2016ECCV}. However, the problem about how to transfer manifold constraints between input data and learned features remains not being well solved.

In this paper, we propose a spatio-temporal manifold \footnote{The spatio-temporal structure is calculated based on a group of manifold sample sets.} network (STMN) approach for action recognition, to alleviate the above problems from the perspective of deep learning regularization. Fig.~\ref{fig.framework} shows the basic idea, where the spatio manifold models the non-linearity of action samples while the temporal manifold considers the dependence structure of action frames. In general, this paper aims to answer ``how the spatio-temporal manifold can be embedded into CNN to improve the action recognition performance".
Specifically, our assumption is that the intrinsic data structure, i.e. manifold structure, can be preserved in the deep learning pipeline, being transferred from the input video sequences into the feature space. With such assumption, CNN is exploited to extract feature maps with respect to the overlapped clips of each video. Meanwhile, a new manifold constraint model is intuitively obtained and embedded into the loss function of CNN to reduce the structure variations in the high-dimensional data. We then solve the resulting constrained optimization problem based on the Alternating Direction Method of Multipliers and Backward Propagation (ADMM-BP) algorithm. In addition, our theoretical analysis shows we can seamlessly fuse the manifold structure constraint with the back propagation procedure through manifold embedding in the feature layer (the last layer of CNN). As a result, we can easily implement our optimization algorithm by additionally using a projection operation to introduce the manifold constraint. \textbf{The main contributions of this paper include}:
	\textbf{(1)} The spatio-temporal manifold is introduced into the loss function of a deep learning model  as a regularization term for action recognition. The resulting STMN reduces the intra-class variations and alleviates the over-fitting problem.	
	\textbf{(2)} A new optimization algorithm ADMM-BP is developed to transfer the manifold structure between the input samples and deep features.

\section{Related Work}\label{Sec.2}

Early methods represent human actions by hand-crafted features~\cite{Dollar2005VSPETS,Laptev2005IJCV,Willems2008ECCV}. Laptevs \etal~\cite{Laptev2005IJCV} proposed space time interest points (STIPs) by extending Harris corner detectors into Harris-3D. SIFT-3D~\cite{Scovanner2007MM} and HOG-3D~\cite{Klaser2008BMVC} descriptors, respectively evolved from SIFT and HOG, were also proposed for action recognition.
Wang \etal~\cite{Wang2013ICCV} proposed an improved dense trajectories (iDT) method, which is the state-of-the-art hand-crafted feature. However, it becomes intractable on large-scale dataset due to its heavy computation cost.

%

		\begin{figure}
			\begin{minipage}[t]{.52\linewidth}
				\centering
			\includegraphics[width=0.97\linewidth]{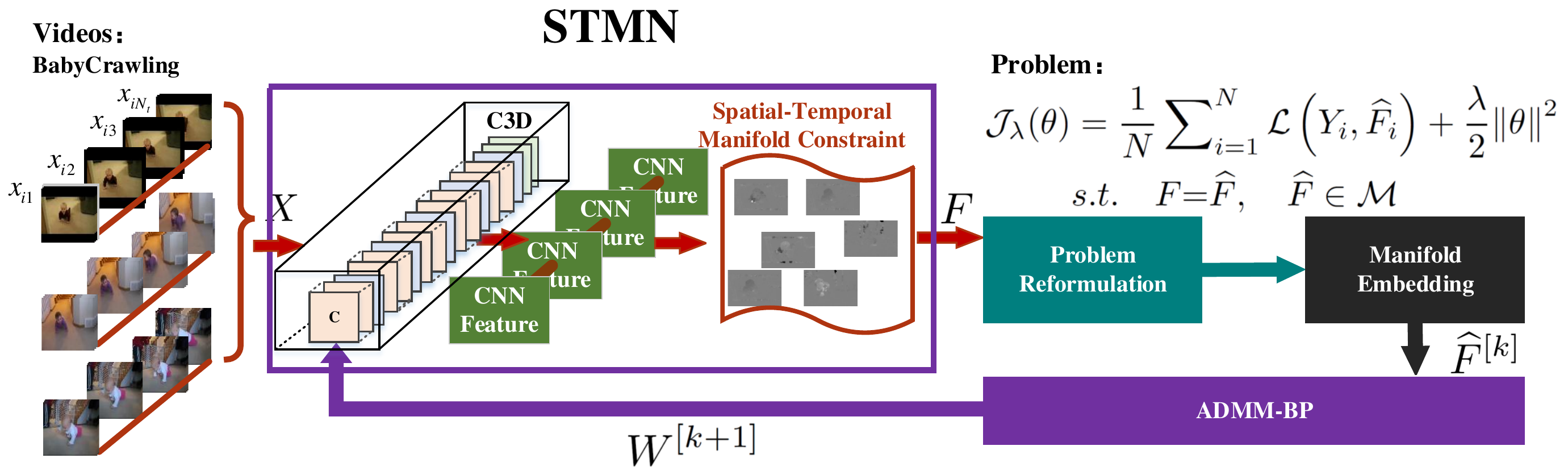}
				\vspace{-0.1cm}
				\caption{The STMN model is solved with an ADMM-BP algorithm, which leads to a chain of compact CNN features for action recognition. STMN is fine-tuned based the C3D model with manifold embedding in the BP procedure.}				
				\label{fig.illustration}
			\end{minipage}
			\hfill
			\begin{minipage}[t]{.44\linewidth}
				\centering
				\includegraphics[width=0.8\linewidth]{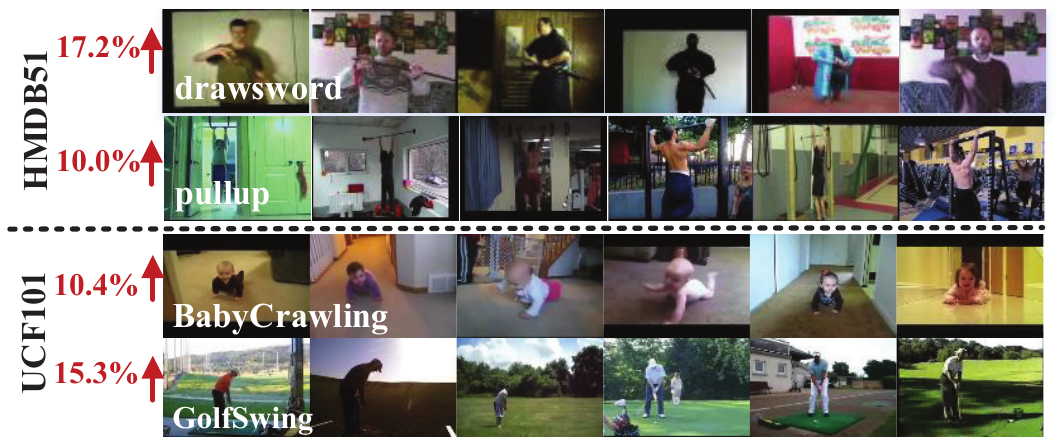}
				\vspace{-0.1cm}
				\caption{Example frames from HMDB51 and UCF101 datasets. (Red numbers indicate the improvements of STMN over C3D for corresponding actions.)}
				
				\label{fig.example}
			\end{minipage}
			\vspace{-0.3cm}
		\end{figure}

Recently, deep learning approaches ~\cite{Huang2012CVPR,Jain2014ICLR} were proposed for human action recognition. Other related works for extracting spatio-temporal features using deep learning include Stacked ISA~\cite{Le2011CVPR}, Gated Restricted Boltzmann~\cite{Taylor2010ECCV} and extended 3D CNN~\cite{Ji2013PAMI}. Karpathy \etal~\cite{Karpathy2014CVPR} trained deep structures on Sports-1M dataset. Simonyan \etal~\cite{Simoyan2014NIPS} designed two-stream CNN containing spatio and temporal nets for capturing motion information. Wang \etal~\cite{Wang2015CVPR} conducted temporal trajectory-constrained pooling (TDD) to aggregate deep convolutional features as a new video representation and achieved state-of-the-art results.

Despite of the good performance achieved by deep CNN methods, they are usually adapted from image-based deep learning approaches, ignoring the time consistency of action video sequences. Tran \etal~\cite{Tran2015CVPR} performed 3D convolutions and 3D pooling to extract a generic video representation by exploiting the temporal information in the deep architecture. Their work called C3D method is conceptually simple and easy to train and use. However, the manifold structure is not explicitly exploited in all existing deep learning approaches.

Our proposed approach builds up the C3D method, and it goes beyond C3D by introducing a new regularization term to exploit the manifold structure during the training process, in order to reduce intra-class variations and alleviate the over-fitting problem. Rather than simply combine the manifold and CNN, we theoretically obtain the updating formula of our CNN model by preserving the structure of the data from the input space to the feature space. Although using the manifold constraint, our work differs from the latest manifold work~\cite{Lu2015CVPR} in the following aspects. First, our method is obtained from a theoretical investigation under the framework of ADMM, while ~\cite{Lu2015CVPR} is empirical. Second, we are inspired from the fact that deep learning is so powerful that it can well discriminate the inter-class samples, and thus only intra-class manifold was considered to tackle the unstructured problem existed in the deep features (in Fig. 1). Differently, the method in ~\cite{Lu2015CVPR} is a little redundant on considering intra-class and inter-class information based on the complicated manifold regularization terms. Our study actually reveals that the inter-class information was already well addressed in deep learning model and there is no need to discuss it again. We target at action recognition, which is not a direct application in ~\cite{Lu2015CVPR}.

\section{Manifold Constrained Network}\label{Sec.3}
We present how the spatio-temporal manifold constraint can be introduced into CNN, i.e. C3D~\cite{Tran2015CVPR}, for action recognition. Fig.~\ref{fig.illustration} shows the framework of STMN, in which the intra-class manifold structure is embedded as a regularization term into the loss function, which eventually leads to a new ADMM-BP learning algorithm to train the CNN model.

We summarizes important variables, wehre  $F$ is the CNN feature map, $\widehat{F}$ is the cloning of $F$, which formulates a manifold.  $\theta$ denotes the weight vector for the last fully connected (FC) layer, and $W$ represents convolution filters for other layers.

\subsection{Problem formulation}\label{Sec.3.1}

Let $X=\left\{X_i\right\}\in{\mathbb{R}},i\in\left[{1,N}\right]$ be a set of $N$ training videos, where  $X_i=\left\{x_{i1},x_{i2},...,x_{iN_t}\right\}$ denotes the $i$th video with $X_i$ divided into $N_t$ clips (see Fig.~\ref{fig.illustration}). $X$ is the input of C3D, and the output feature map is denoted as $F$. Given the convolution operator $\odot$ and max pooling operator $\Psi$, the network performs convolutions in the spatio-temporal domain with a number of filter weights $W$ and bias $b$. The function in the convolution layer is ${f_W}(X_i) = \Psi ({W \odot X_i + b})$. In the last FC layer, the loss function for the $L$-layers network is:
\begin{equation}
{{\cal J}_{\lambda}}(\theta) = \frac{1}{N}\sum\nolimits_{i = 1}^N {{\cal L}\left(Y_i,{f_W}(X_i)\right)}  + \frac{\lambda }{2}{{{\lVert {{\theta}} \rVert}^2}},
\label{eq.2}
\end{equation}
where $\theta$ denotes the weight vector in the last FC layer, and all biases are omitted. In Eq.~(\ref{eq.2}), the softmax loss term ${{\cal L}\left(Y_i,{f_W}(X_i)\right)}$ is
\begin{equation}
{{\cal L}\left(Y_i,{f_W}(X_i)\right)=-\log \frac{{{e^{\theta_{{Y_i}}^{T}{f_W}(X_i)+{b_{Y_i}}}}}}{{\sum\nolimits_{j = 1}^m {{e^{\theta_j^{T}{f_W}(X_i)+{b_j}}}} }}},
\label{eq.lossterm}
\end{equation}
where ${f_W}(X_i)$ denotes the deep feature for $X_i$, belonging to the $Y_i$th class. $\theta_j$ denotes the $j$th column of weights in the last FC layer, $m$ is the number of classes. To simplify the notation, we denote the output feature map for video $X_i$ as $F_i=\left\{F_{i1}^L,F_{i2}^L,...,F_{iN_t}^L\right\}$, which is able to describe the nonlinear dependency of all features $F_{ij}^L$ after $L$ layers for video clips. As a result, the deep features are denoted as $F=\left\{F_i\right\},i\in\left[{1,N}\right]$, and $F^{[k]}$ refers to the learned feature at the $k$th iteration (see Fig.~\ref{fig.illustration}).

The conventional objective function in Eq.~(\ref{eq.2}) overlooks a property that the action video sequences usually formulate a specific manifold ${\cal M}$, which represents the nonlinear dependency of input videos. For example, in Fig.~\ref{fig.framework} the intra-class of video $X$ with separated clips lies on a spatio-temporal manifold ${{\cal M}}$, which is supported by the evidence that video sequence with continuously moving and/or acting objects often lies on specific manifolds~\cite{Zhang2015CVPR}.
To take advantage of the property that the structure of the data can actually contribute to better solutions for lots of existing problems~\cite{Lu2015CVPR}, we deploy a variable cloning technique $X = \widehat X$ with $ \widehat X \in {\cal M}$ to explicitly add manifold constraint into the optimization objective Eq.~(\ref{eq.2}). We then have a new problem:
\begin{equation*}
\begin{aligned}
{{\cal J}_{\lambda}}(\theta)&= \frac{1}{N}\sum\nolimits_{i = 1}^N {{\cal L}\left(Y_i,{f_W}( \widehat X_i)\right)}  + \frac{\lambda }{2} {{{\lVert {{\theta}} \rVert}^2}} \quad
s.t. \quad X{\rm{ = }}\widehat X,\quad \widehat X \in {\cal M}.
\end{aligned}
\tag{P1}
\label{P1}
\end{equation*}
(\ref{P1}) is more reasonable since the intrinsic structure is considered.
However, it is unsolvable because $\theta$ is for the last FC layer of CNN and is not directly related to the input $X$.

In the deep learning approach with error propagation from the top layer, it is more favorable to impose the manifold constraint on the deep layer features. This is also inspired from the idea of manifold on the structure for preserving in different spaces, i.e. the high-dimensional and the low-dimensional spaces. Similarly, the manifold structure of $X$ in the input space is assumed to be preserved in the feature $F$ of CNN in order to reduce variation in the higher-dimensional feature space (Fig.~\ref{fig.framework}). That is to say, an alternative manifold constraint is obtained as $ F \in {\cal M}$, and evidently $F$ is more related to CNN training. To use $ F \in {\cal M}$ to solve the problem~(\ref{P1}), we perform variable replacement, i.e. $F = \widehat F$, alternatively formulate $\widehat F$ as a manifold, and achieve a new problem~(\ref{P2}),
\begin{equation*}
\begin{aligned}
{{\cal J}_{\lambda}}(\theta)&= \frac{1}{N}\sum\nolimits_{i = 1}^N {{\cal L}\left(Y_i,\widehat F_i\right)}  + \frac{\lambda }{2}{{{\lVert {{\theta}} \rVert}^2}} \quad
s.t. \quad F{\rm{ = }}\widehat F,\quad \widehat F \in {\cal M}.
\end{aligned}
\tag{P2}
\label{P2}
\end{equation*}
It is obvious that the objective shown in problem~(\ref{P2}) is learnable, because $F$ is the convolution result based on the learned filter ($W$ and $\theta$) and $\theta$ is directly related to $F$.

\subsection{ADMM-BP solution~(\ref{P2})}\label{Sec.3.2}

Based on the augmented lagrangian multiplier (ALM) method, we have a new objective for the problem~(\ref{P2}) as
\begin{equation}
\begin{aligned}
{{\cal J}_{\lambda ,\sigma }}( {\widehat F}, F; \theta, R) = {{\cal J}_\lambda }(\theta) + {R^T}( {\widehat F - F} ) + \frac{\sigma }{2}{\lVert {\widehat F - F} \rVert^2},
\label{eq.3}
\end{aligned}
\end{equation}
where ${R^T}$ denotes the Lagrange multiplier vector, $\sigma $ is the corresponding regularization factor. Optimizing the above objective involves complex neural network training problem. Eq.~(\ref{eq.3}) is solved based on ADMM and backward propagation algorithm, named ADMM-BP, which integrates CNN training with manifold embedding in an unified framework.

Specifically, we solve each variable in each sub-problem. ADMM-BP is described from the $k$th iteration, and ${\widehat F^{[k]}}$ is first solved based on ${ F^{[k]}}$. Next ${ F^{[k]}}$, $R^{[k+1]}$, $\theta^{[k+1]}$ and $W^{[k+1]}$ are solved step by step. Finally ${ F^{[k+1]}}$ is obtained, which is then used to calculate ${\widehat F^{[k+1]}}$ similar to that in the $k$th iteration. We have
\begin{equation}
\begin{aligned}
{{\widehat F}^{[k]}} &= \arg \min{{\cal J}_{\lambda ,\sigma }}( {\widehat F|{F^{[k]}}} )\quad
s.t.\quad \widehat F \in {\cal M},
\end{aligned}
\label{eq.6}
\end{equation}
which is described in the next section. And then
\begin{equation}
R^{[k+1]} = R^{[k]} + \sigma^{[k]}( \widehat{F}^{[k]} - F^{[k]} ).
\label{eq.61}
\end{equation}
For the FC layer, we use the gradient descend method,
\begin{equation}
\small
\theta^{[k+1]} = \theta ^{[k]}  - \alpha \frac{{\partial {{\cal J}_{\lambda ,\sigma }^{[k]}}}}{{\partial {\theta^{[k]}}}}=\theta^{[k]}  - \alpha \frac{{\partial {{\cal J}_{\lambda }^{[k]}}}}{{\partial {\theta^{[k]}}}},
\label{eq.ll}
\end{equation}
and we update the parameters for  convolutional layers $W$ by stochastic gradient descent in the backward propagation as
\begin{equation}
\small
{W^{[k + 1]}}= {W^{[k]}} - \alpha \frac{{\partial {{\cal J}_{\lambda ,\sigma }^{[k]}}}}{{\partial \widehat F^{[k]}}}\cdot \frac{{\partial \widehat F^{[k]}}}{{\partial {W^{[k]}}}},\\
\label{eq.17}
\end{equation}
where $\alpha$ is the learning rate, $k$ is the iterative number, and
\begin{equation}
\small
\frac{{\partial {{\cal J}_{\lambda ,\sigma }^{[k]}}}}{{\partial \widehat F^{[k]}}} = \frac{{\partial {{\cal J}_{\lambda}^{[k]} }}}{{\partial \widehat F^{[k]}}} + \sigma^{[k]} \left( {{\widehat F^{[k]}} - F^{[k]}} \right) + {R^{[k]T}}.
\label{eq.lice}
\end{equation}
Now we have an updated CNN model to calculate the feature map ${ F^{[k+1]}}$, which is then deployed to calculate ${\widehat F^{[k+1]}}$ via Eq.~(\ref{eq.6}) (replacing $k$ by $k+1$).
\subsection{Manifold embedding}\label{Sec.3.3}
In the ADMM-BP algorithm, only Eq.~(\ref{eq.6}) is unsolved because of an unknown manifold constraint $\cal{M}$. Based on Eq.~(\ref{eq.3}), we can rewrite Eq.~(\ref{eq.6}) by dropping the constant terms and the index of variables,
\begin{equation}
\begin{aligned}
\widehat F &= \arg \min [ {{R^T}( {\widehat F - F} ) + \frac{\sigma }{2}{{\lVert {\widehat F- F} \rVert}^2}}]
= \arg \min {\lVert {\widehat F - ( {F - \frac{R}{\sigma }})} \rVert^2}\quad
s.t.\quad \widehat F \in {\cal M}.
\end{aligned}
\label{eq.8}
\end{equation}

In the $k$th iteration, we have $\widehat F^{[k]}{\rm{ = }}{A_{\cal M}}( {F^{[k]} - \frac{R^{[k]}}{\sigma^{[k]} }} )$\footnote{We have $\widehat F^{}{\rm{ = }} ( {F^{} - \frac{R^{}}{\sigma^{} }} )$ without manifold constraint.}, where $A_{\cal M}$ is the projection matrix related to the manifold $\cal M$. This is the key part of the proposed algorithm where the constraint manifold ${\cal M}$ arises. Replacing ${\cal M}$ equals replacing the projection $A_{\cal M}$. This is the modularity which we alluded previously. To calculate $A_{\cal M}$, we exploit the Locally Linear Embedding (LLE) method~\cite{Roweis2000Science} in order to find a structure-preserving solution for our problem based on the embedding technique.  By considering intrinsic manifold structure of the input data, the algorithm can stop on a manifold,  $A_{{\cal M}}$, in the $k$th iteration as
\begin{equation}
A_{{\cal M}} ={F}_{[1:H]} \Omega^{[k]},
\label{eq.10}
\end{equation}
where $\Omega^{[k]}$ is a diagonal matrix defined as  $\Omega^{[k]} = {\rm{diag}}({\omega}^{[k]}_1,...,{\omega}^{[k]}_N)$. ${F}_{[i1:iH]}$ are the $H$ neighborhoods of the sample and ${\omega}^{[k]}_N$ are the corresponding weight vector calculated in LLE.

{\bf Algorithm.} The ADMM-BP algorithm is summarized in Alg.~\ref{alg.admm}, where the key step defined by Eq. (12) is respectively solved in Sec.~\ref{Sec.3.2} and Sec.~\ref{Sec.3.3}. Although the convergence of the ADMM optimization problem with multiple variables remains an open problem, our learning procedures experimentally never diverge, because new adding variables related to manifold constraint are solved following the similar pipeline of back propagation. Based on the learned STMN model, we obtain a chain of CNN features denoted as
\begin{equation}
\begin{aligned}
\widetilde F&=\left\{\widetilde F_i\right\}\in{\mathbb{R}}, i\in\left[{1,N}\right], \quad
\widetilde{F}_i&=\left\{\widetilde F_{i1},\widetilde F_{i2},...,\widetilde F_{iN_t}\right\},
\end{aligned}
\end{equation}
where $N$ is the number of videos, and $\widetilde{F}_i$ is the STMN feature for the video $X_i$  with $N_t$ clips.
\begin{algorithm}[htb]
	\label{alg.admm}
	\caption{\textbf{ADMM-BP for the problem~(\ref{P2})}}
	\begin{algorithmic}[1]
		\small
		\STATE Set $t = 0$ and $\epsilon_{\sf best} = +\infty$
		\STATE Initialize $\alpha$, $\lambda$, $\sigma^{[0]}$, $R^{[0]}$, and $0 < \eta <= 1$
		\STATE Initialize $\theta^{[0]}$, $W^{[0]}$, ${\widehat{F}}^{[0]}$, and ${\Omega^{[0]}}$
		\REPEAT
		\STATE
		\begin{equation}
		\hspace{-6em}
		({\widehat{F}}^{[k+1]},R^{[k+1]},\theta^{[k+1]},W^{[k+1]}) = \begin{array}[t]{ll} \arg \min {\cal J}_{\lambda ,\sigma^{[k]}}(\widehat{F}, F; \theta^{[k]}, R^{[k]} | {\Omega}^{[k]})
		\;\;~\mbox{\textit{s.t.}} \widehat{F} \in {\mathcal M} 
		\end{array}
		\end{equation}
		\\
		Update ${\Omega}^{[k]}$ and ${\widehat F}^{[k]}$ by LLE\hspace*{8ex}
		\STATE \hspace*{4ex}  $\epsilon = \lVert{\widehat{F}}^{[k+1]} - {\widehat{F}}^{[k]}\rVert^2$
		\STATE \hspace*{4ex} \textbf{if} $\epsilon < \eta\, \epsilon_{\sf best}$
		\STATE \hspace*{8ex} $R^{[k+1]} = R^{[k]} + \sigma^{[k]} ( \widehat{F}^{[k]} - F^{[k]} )$
		\STATE \hspace*{8ex} $\sigma^{[k+1]} = \sigma^{[k]}$
		\STATE \hspace*{8ex} $\epsilon_{\sf best} = \epsilon$
		\item \hspace*{4ex} \textbf{else}
		\item \hspace*{8ex} $R^{[k+1]} = R^{[k]}$
		\STATE \hspace*{8ex} $\sigma^{[k+1]} = 2 \cdot \sigma^{[k]}$
		\STATE \hspace*{4ex} \textbf{endif}
		\STATE \hspace*{4ex} $k \leftarrow k+1$
		\UNTIL \textit{maximum iteration step or $\epsilon \le 0.001$}
	\end{algorithmic}
\end{algorithm}
%
\section{Experimental Setting and Results}\label{Sec.4}

Two benchmark datasets are used to validate our approach for action recognition.

{\bf HMDB51~\cite{HMDB51}} consists of  $6,766$ realistic videos from $51$  action categories with each category containing at least 100 videos. We follow the evaluation scheme in~\cite{HMDB51} to report the average accuracy over three different training/testing splits.

{\bf UCF101~\cite{UCF101}} contains $101$ action classes with each class having at least $100$ videos. The whole dataset contains $13,320$ videos, which are divided into $25$ groups for each action class. We follow the evaluation scheme of the THUMOS13 Challenge~\cite{THUMOS13} to use the three training/testing splits for performance evaluation. Example frames from both datasets are shown in Fig.~\ref{fig.example}.


{\bf Learning strategies.}
We initially use UCF101 dataset to train the STMN network, which is further deployed for feature extraction on all datasets. We use the  C3D~\cite{Tran2015CVPR}, which is a 3D version of CNN designed to extract temporal features to obtain a chain of CNN features for video recognition. In C3D, each video is divided into 16-frame clips with 8-frame overlapped between two consecutive clips as the input of the network. The frame resolution is set to  $128\times171$, and input sizes of C3D are $3\times16\times128\times171$ (channels$\times$frames$\times$height$\times$width). The C3D network uses 5 convolution layers, 5 pooling layer, 2 FC layers and a softmax loss layer to predict action labels. The filter numbers from the first to fifth convolutional layer respectively are 64, 128, 256, 256 and 256. The sizes of convolution filter kernels and the pooling layers respectively are  $3\times3\times3$ and  $2\times2\times2$. The output feature size of each FC layer is  4096.  The proposed STMN is trained using mini-batch size of 50 examples with initial learning rate $\lambda = 0.001$. The resulting network is further used to extract the 4096-dim features for each video clip. All clips are finally concatenated as the features.

{\bf Classification model and baseline.}
We train and test STMN features ($\widetilde F$) using the multi-class linear SVM. More specifically, STMN was trained on the split1 of UCF101, and the learned network is then used to extract features on both HMDB51 and UCF101. In the testing stage, we followed the same protocol as used in TDD~\cite{Wang2015CVPR}, TSN~\cite{Wang2016ECCV} and C3D~\cite{Tran2015CVPR}.
Both the state-of-the-art handcrafted and deep learning methods including DT+BoVW~\cite{Wang2013IJCV}, DT+MVSV~\cite{Cai2014CVPR}, iDT+FV~\cite{Wang2013ICCV}, DeepNet~\cite{Karpathy2014CVPR}, Two-stream CNN~\cite{Simoyan2014NIPS}, TDD~\cite{Wang2015CVPR}, TSN~\cite{Wang2016ECCV} and C3D~\cite{Tran2015CVPR}, are employed for an extensive comparison. C3D is used as the baseline.

\subsection{Results and analysis}\label{Sec.4.3}
\begin{figure*}[htb]
	\begin{center}
		\begin{minipage}{0.21\linewidth}
			\centering
			\includegraphics[width=\linewidth,height=2.8cm]{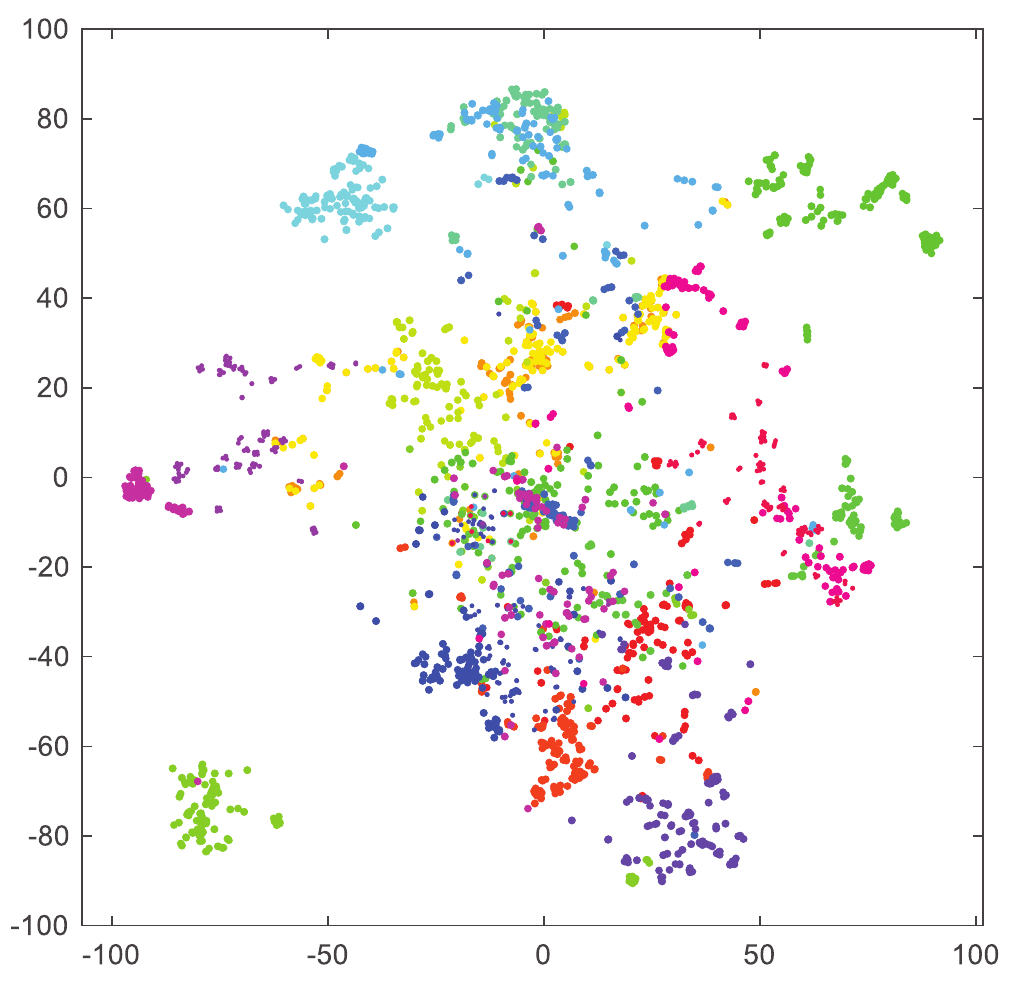}
			\centerline{(a)}
		\end{minipage}
		\begin{minipage}{0.27\linewidth}
			\centering
			\includegraphics[width=\linewidth,height=2.8cm]{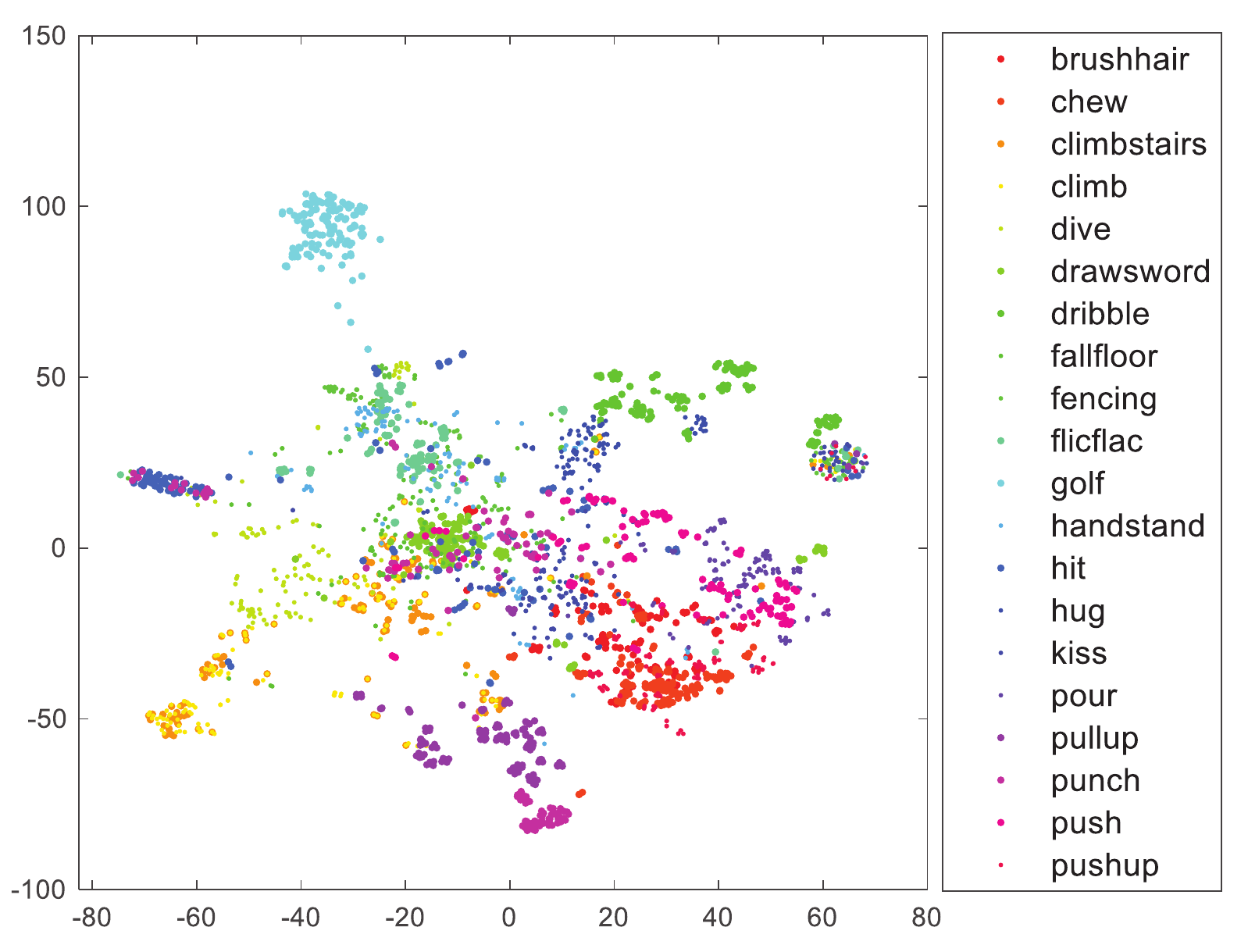}
			\centerline{(b)}
		\end{minipage}
		\begin{minipage}{0.21\linewidth}
			\centering
			\includegraphics[width=\linewidth,height=2.8cm]{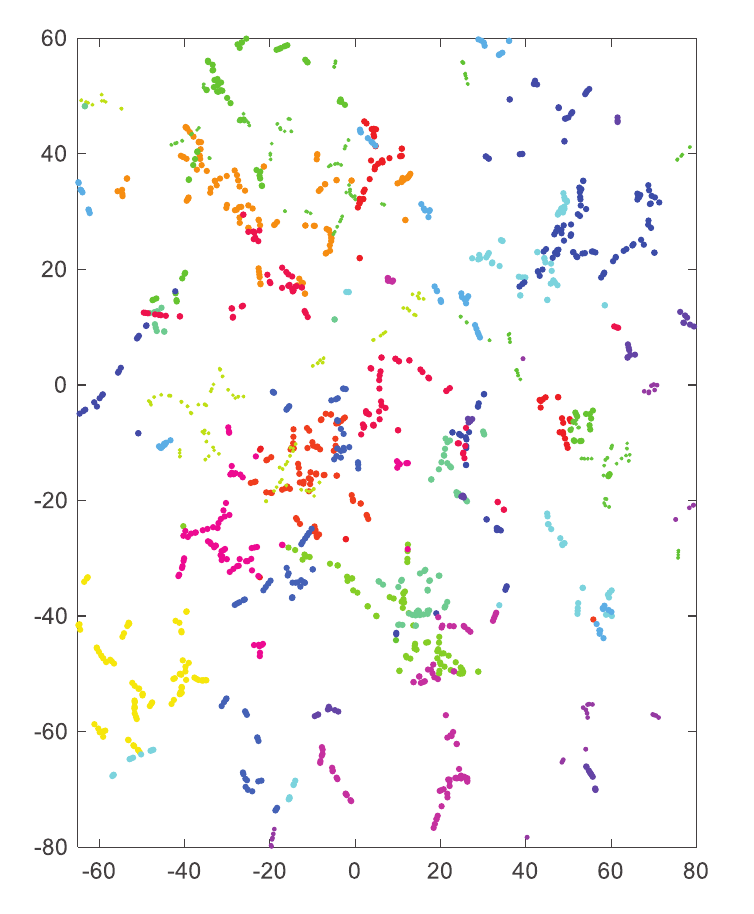}
			\centerline{(c)}
		\end{minipage}
		\begin{minipage}{0.27\linewidth}
			\centering
			\includegraphics[width=\linewidth,height=2.8cm]{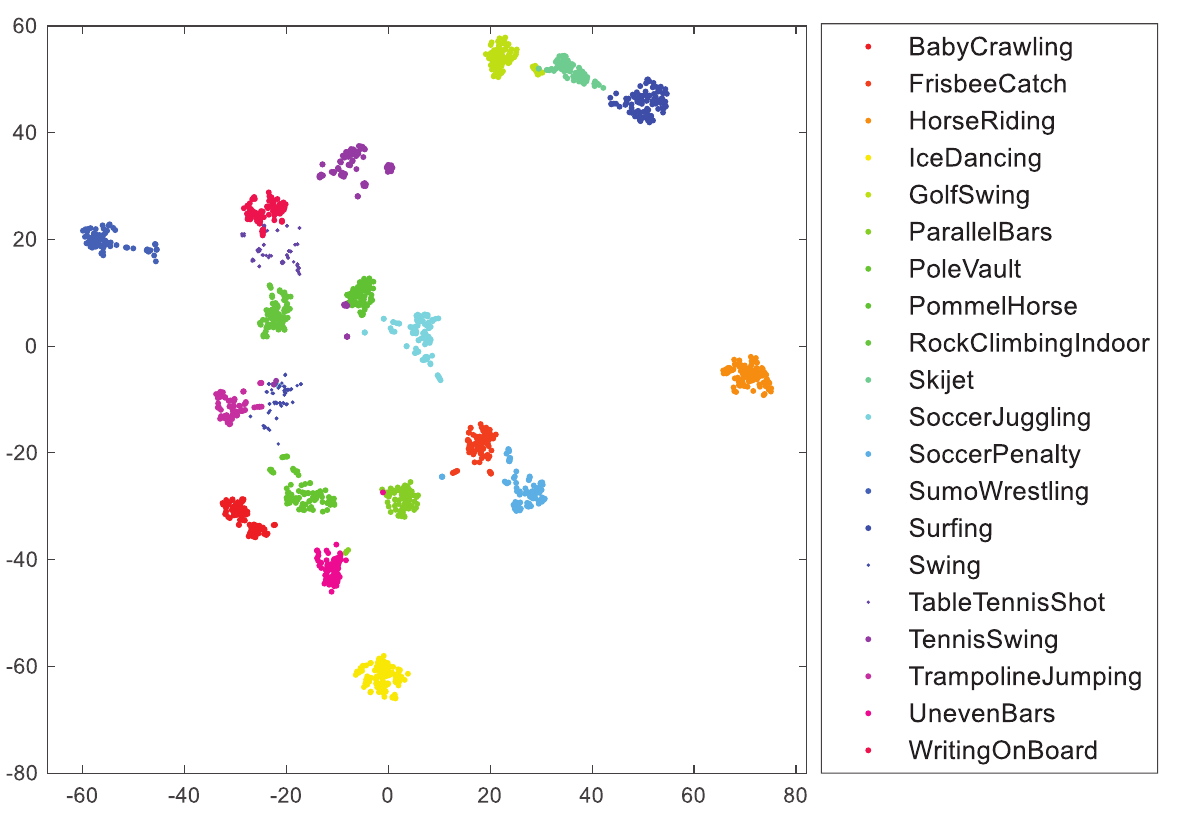}
			\centerline{(d)}
		\end{minipage}
	\end{center}
	\caption{Feature visualization of twenty difficult classes. (a) and (c) are C3D features on the HMDB51 and UCF101 dataset;  (b) and (d) are STMN features on the two datasets. The STMN feature is more discriminative than the C3D feature.}
	\label{fig.embedVisual}
\end{figure*}


		\begin{figure}
			\begin{minipage}[t]{.47\linewidth}
				\centering
				\includegraphics[width=0.88\linewidth]{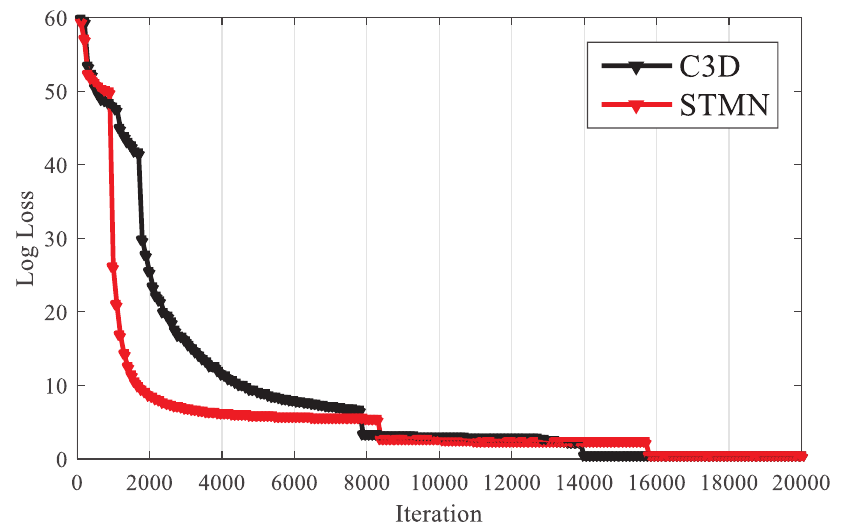}
				\vspace{-0.1cm}
				\caption{Convergence analysis on the split1 of UCF101.}
				\vspace{-0.2cm}
				\label{fig.convCurve}
			\end{minipage}
			\hfill
			\begin{minipage}[t]{.47\linewidth}
				\centering
				\includegraphics[width=0.95\linewidth]{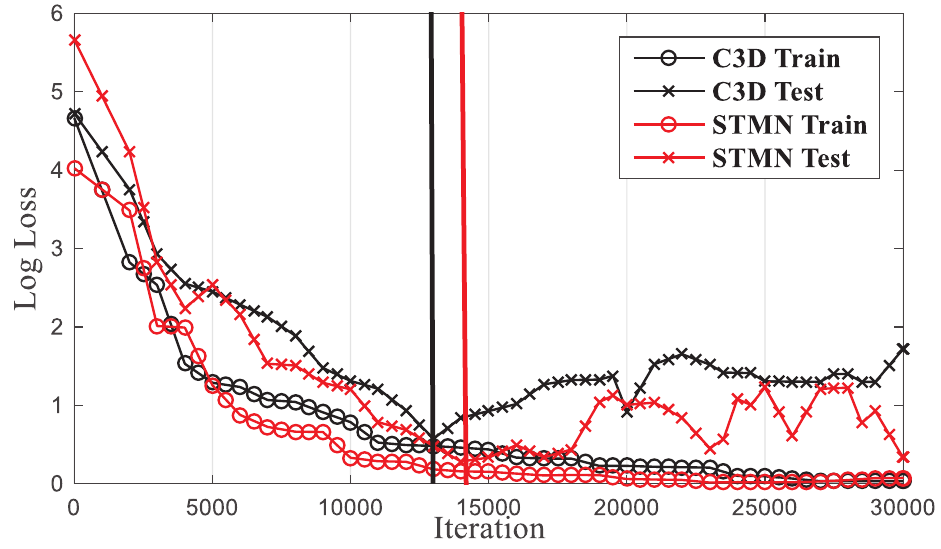}
				\vspace{-0.1cm}
				\caption{Over-fitting validation on different iterations.}
				\vspace{-0.2cm}
				\label{fig.overfit}
			\end{minipage}
		\end{figure}

We first study the average recognition accuracies of our STMN when using different number of neighborhoods in LLE in Table.~\ref{table.neigh}. Due to limited number of training videos in each class, we learned the STMN on the split1 in UCF101 using $H=5,15,20$ neighbor samples and extracted features. STMN achieves the best accuracies of $69.7\%$ and $92.5\%$ on HMDB51 and UCF101 respectively, when $H=20$. Note that the value of $H$ has to be smaller than the
batch size. In our experiment, we can only evaluate the performance of our STMN by setting $H$ up to 20 due to memory limitation of GPUs.

Fig.~\ref{fig.embedVisual} shows the embedding feature visualizations on HMDB51 and UCF101 datasets. In Fig.~\ref{fig.embedVisual}(a) and Fig.~\ref{fig.embedVisual}(c), the C3D features of twenty difficult classes on HMDB51 and UCF101 are visualized by t-SNE~\cite{Maaten2009Science}, while the STMN features are illustrated in Fig.~\ref{fig.embedVisual}(b) and Fig.~\ref{fig.embedVisual}(d), respectively. Clearly, the STMN feature is more discriminative than the C3D feature, especially the STMN feature in Fig.~\ref{fig.embedVisual}(d) can be better discriminated  than the C3D feature in Fig.~\ref{fig.embedVisual}(b). As another verification, the quantitative evaluation is performed based on intra-class mean and variance in the next section.

\begin{figure}
	\begin{center}
		\begin{minipage}{0.48\linewidth}
			\centering
			\includegraphics[width=0.93\linewidth,height=3.4cm]{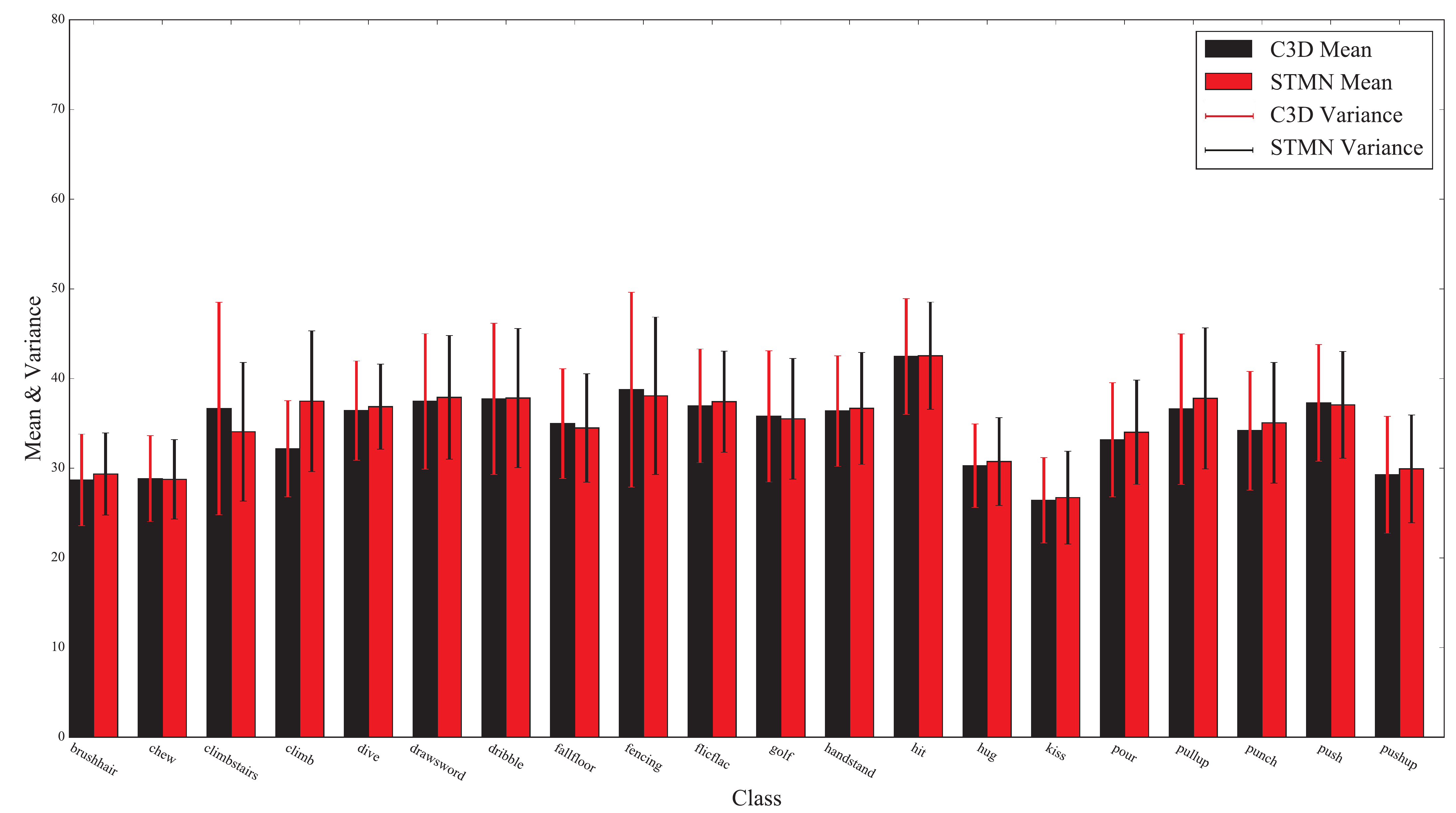}
			\centerline{(a)}
		\end{minipage}
		\begin{minipage}{0.48\linewidth}
			\centering
			\includegraphics[width=0.93\linewidth,height=3.4cm]{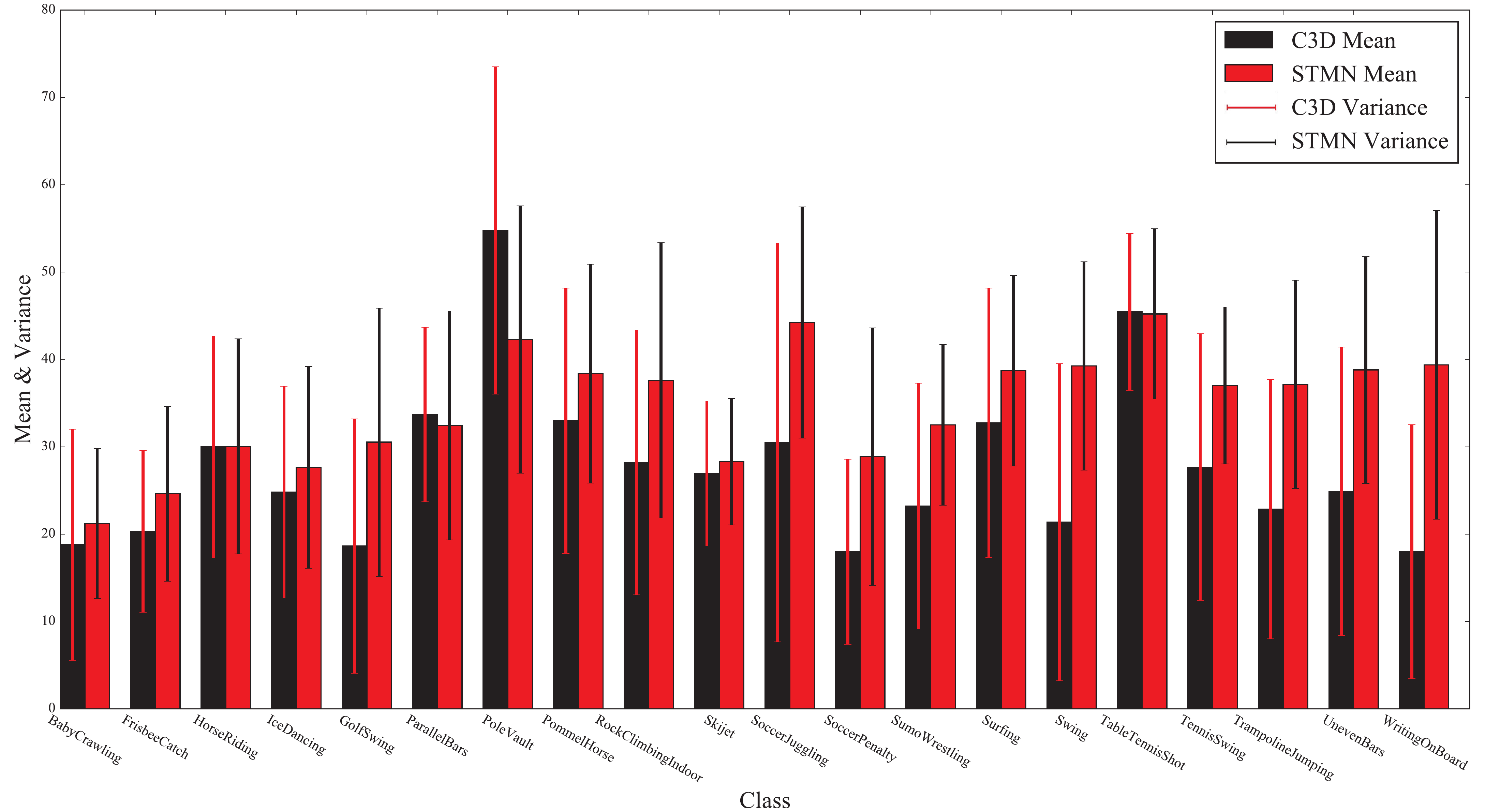}
			\centerline{(b)}
		\end{minipage}
	\end{center}
	\vspace{-1em}
	\caption{Comparison of intra-class means and variances of features extracted by using C3D and STMN. The action classes are shown in Fig.~\ref{fig.embedVisual} from (a) HMDB51 and (b) UCF101.  }
	\label{fig.visualmv}
\end{figure}


\begin{table}
\parbox{.4\linewidth}{
\centering
	\begin{center}
	\footnotesize
		\begin{tabular}{|c|c|c|}
			\hline
			{Neighborhoods \#}  & HMDB51 & UCF101 \\
			\hline
			\hline
			$H=5$ &$68.2$  &$75.0$  \\
			$H=15$ &$68.7$  &$86.1$ \\
			$H=20$ & $69.7$  & $92.5$ \\
			\hline
		\end{tabular}
	\end{center}
	\caption{Accuracy (\%) vs. different neighborhoods as manifold constraints for STMN.}
	\label{table.neigh}
}
\hfill
\parbox{.5\linewidth}{
\centering
	\begin{center}
	\scriptsize
		\begin{tabular}{|l|c|c|c|}
			\hline
			Method & HMDB51 & UCF101  & Year\\
			\hline
			\hline
			DT+BoVW & $46.6$ & $79.9$ & $2013$\\
			DT+MVSV& $55.9$ & $83.5$ & $2014$\\
			iDT+FV  & $57.2$ & $84.7$ & $2013$\\
			\hline
			DeepNet&--& $63.3$ & $2014$\\
			Two-stream CNN& $59.4$ & $88.0$ & $2014$\\
			TDD  & $63.2$ & $90.3$ &  $2015$\\
			TSN &$69.4$ &$94.2$ & $2016$\\
			\hline
			C3D (baseline) & $68.4$ & $85.2$  & $2015$\\
			STMN & $\bf{69.7}$ & $\bf{92.5}$ & \\
			\hline
		\end{tabular}
	\end{center}
	\caption{Average recognition accuracies (\%) of different methods.}
	\label{table.accuracy}
}
\end{table}

{\bf Model effect:}
\textbf{1) Convergence:} We employed the parallel computing strategy to utilize GPUs during the training process, which is implemented with our modified version of Caffe. Our STMN is trained on the UCF101 database, which takes about 2 days with two K80 GPUs and Xeon(R) E5-2620 V2 CPU. We plotted the training loss of two algorithms in Fig.~\ref{fig.convCurve}. It is clear that our STMN (the red line) converges much faster than C3D. \textbf{2) Intra-class variation:} As shown in Fig.~\ref{fig.embedVisual} and Fig.~\ref{fig.visualmv}, STMN can exploit the manifold structure to better eliminate randomness of samples in the feature space. Especially as shown in Fig.~\ref{fig.visualmv}, the quantitative intra-class means and variances\footnote{The statistics are computed using pairwise Euclidean distance.} of our STMN features are much smaller than those of C3D, e.g. the total mean on the UCF dataset has decreased from $14.02$ to $11.15$. We can also observe that $21.22$ (STMN mean) versus $18.78$ (C3D mean) and $8.58$ (STMN variance) versus $13.23$ (C3D variance) for the specific action \emph{BabyCrawling}. \textbf{3) Over-fitting:} The manifold regularization can mitigate the over-fitting problem especially when there are not enough training samples in practical applications. In Fig.~\ref{fig.overfit}, we conducted a training experiment using 70 percent of training and testing data from the UCF101 dataset (split2), which shows that C3D overfits the training data at the $12500$th iteration, while our STMN overfits the training data at the $14500$th iteration.


\begin{figure}[htbp]
	\begin{center}
		\begin{minipage}{0.23\linewidth}
			\centering
			\includegraphics[width=0.9\linewidth,height=1cm]{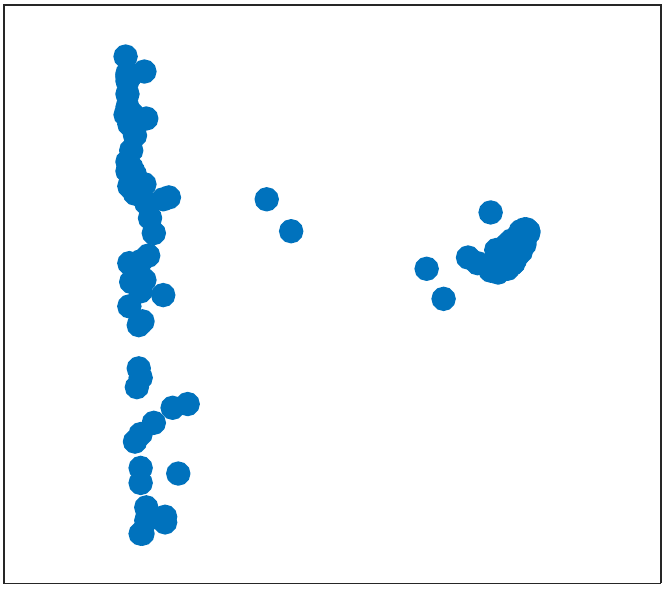}\\
			\includegraphics[width=0.9\linewidth,height=1cm]{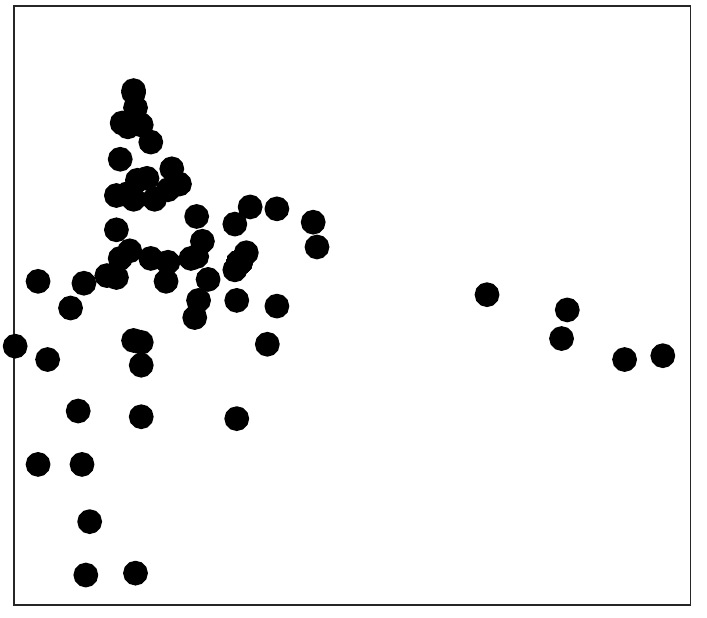}\\
			\includegraphics[width=0.9\linewidth,height=1cm]{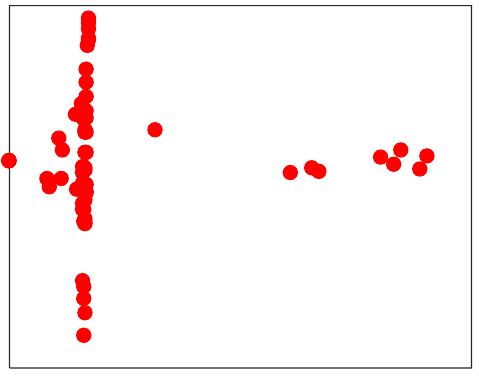}
			\centerline{\emph{drawsword}}
		\end{minipage}
		\begin{minipage}{0.23\linewidth}
			\centering
			\includegraphics[width=0.9\linewidth,height=1cm]{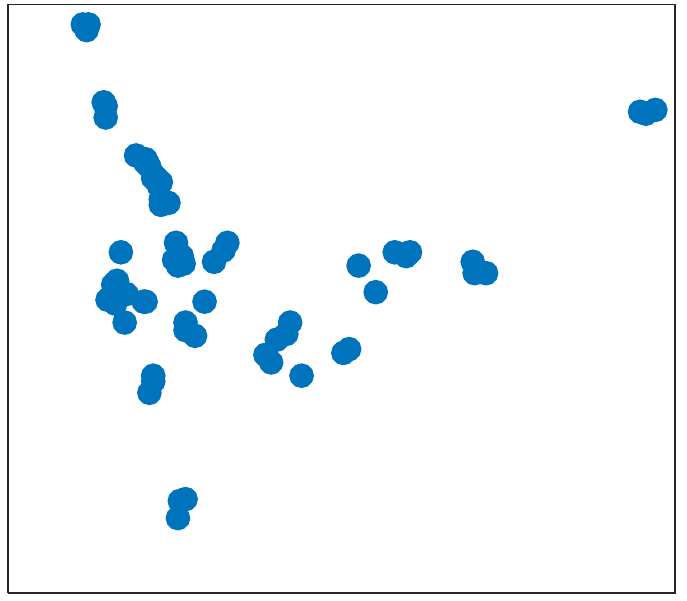}\\
			\includegraphics[width=0.9\linewidth,height=1cm]{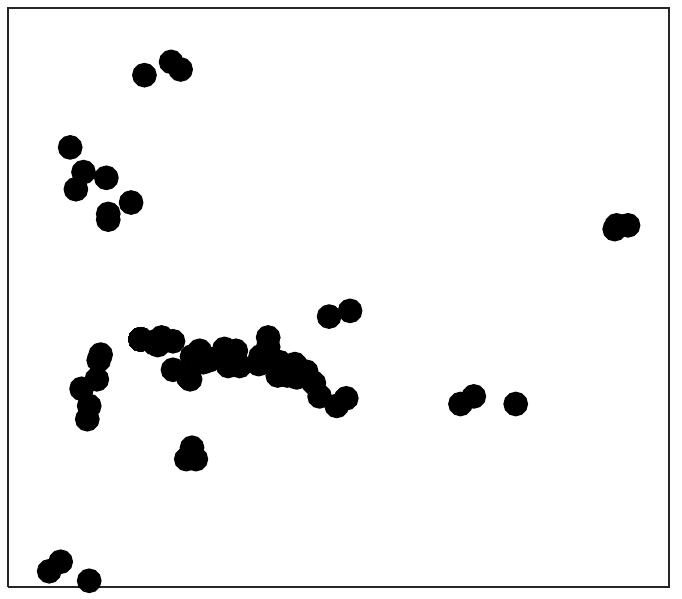}\\
			\includegraphics[width=0.9\linewidth,height=1cm]{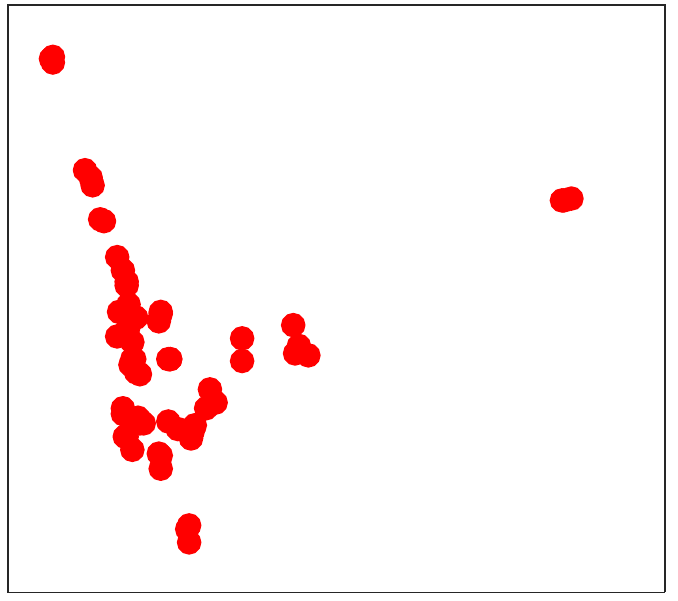}
			\centerline{\emph{pullup}}
		\end{minipage}
		\begin{minipage}{0.23\linewidth}
			\centering
			\includegraphics[width=0.9\linewidth,height=1cm]{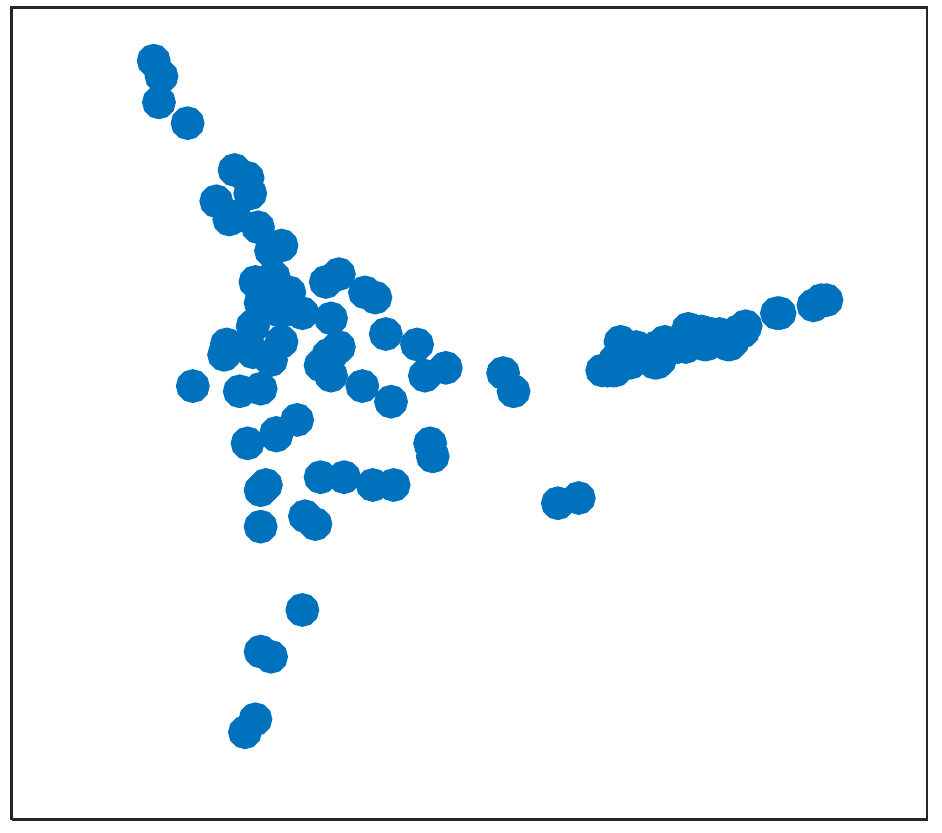}\\
			\includegraphics[width=0.9\linewidth,height=1cm]{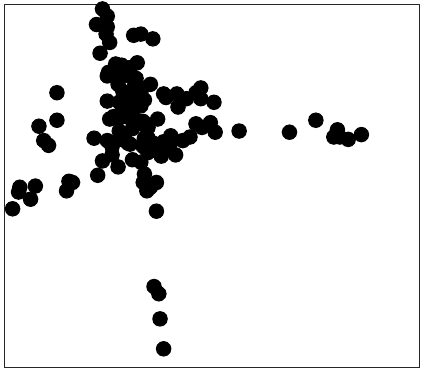}\\
			\includegraphics[width=0.9\linewidth,height=1cm]{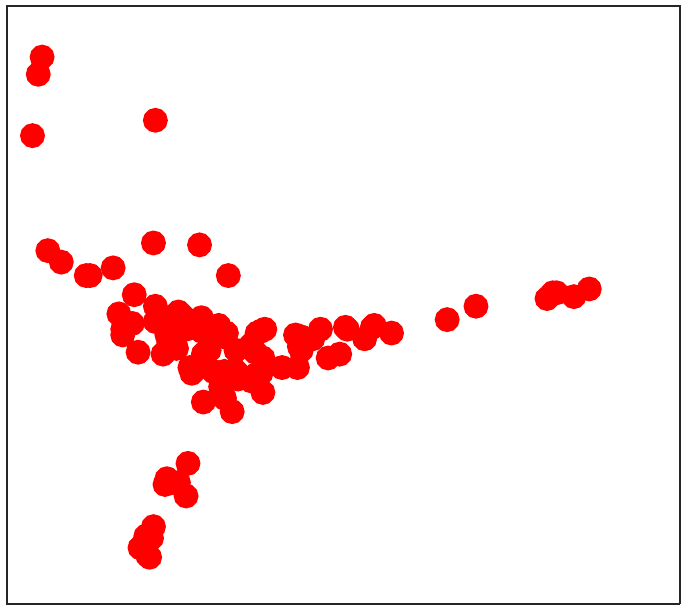}
			\centerline{\emph{BabyCrawling}}
		\end{minipage}
		\begin{minipage}{0.23\linewidth}
			\centering
			\includegraphics[width=0.9\linewidth,height=1cm]{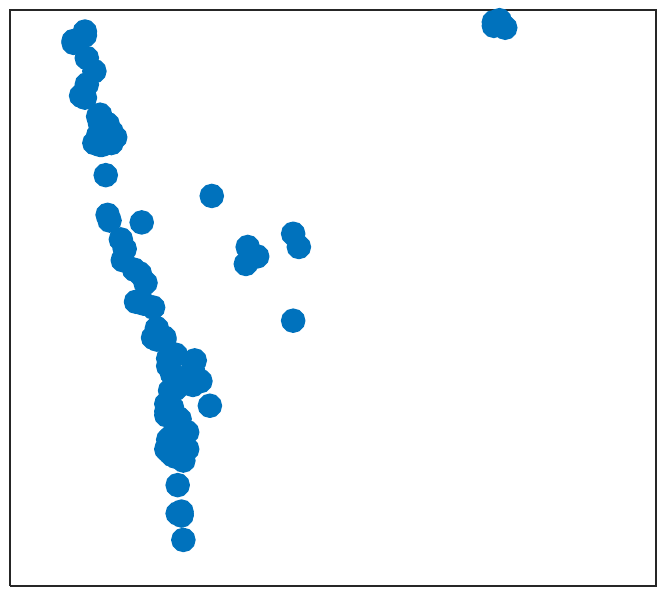}\\
			\includegraphics[width=0.9\linewidth,height=1cm]{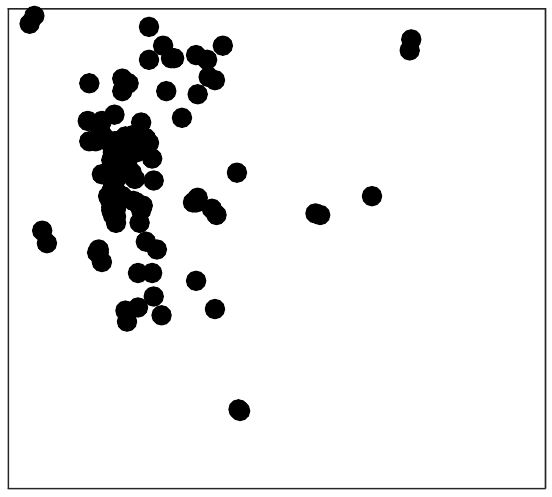}\\
			\includegraphics[width=0.9\linewidth,height=1cm]{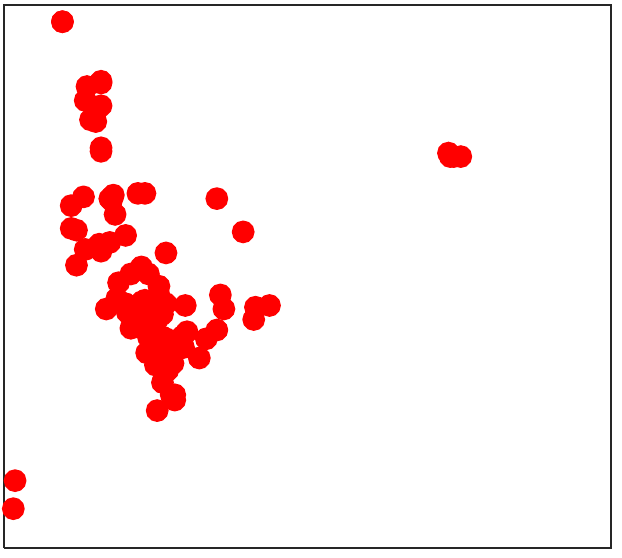}
			\centerline{\emph{GolfSwing}}
		\end{minipage}
	\end{center}
	\caption{Manifold structure visualization (t-SNE) of input data (blue), C3D features (black) and STMN (red) for the examples in Fig.~\ref{fig.example}. }
	\label{fig.msv}
\end{figure}

{\bf Comparisons.} 
We follow the same evaluation scheme to compare our STMN with several representative action recognition methods. The results are shown in Tab.~\ref{table.accuracy}.  STMN also achieves much better results than TDD, which combines the hand-crafted features and deep learning features.  Our STMN also outperforms TSN~\cite{Wang2016ECCV} on the HMDB51 dataset and achieves comparable results on the  UCF101 dataset. It is worth mentioning that TSN is designed based on three modalities of features (RGB, Optical Flow and Warped Flow) through the two-stream deep CNN learning framework, while we only use the RGB features in our STMN and it does not introduce any preprocessing steps used such as extraction of  optical flow and warped flow as in TSN and traditional hand-crafted features based approaches.

In Fig.~\ref{fig.example}, we take four classes including \emph{drawsword, pullup, BabyCrawling, and GolfSwing} as examples for more detailed analysis. The recognition accuracies of our STMN for these four actions are $100\%$, $98\%$, $99\%$ and $100\%$, and the improvements over C3D are $17.2\%$, $10.0\%$, $10.4\%$ and $15.3\%$, respectively. In Fig.~\ref{fig.msv}, we plot the input data, C3D features, and our STMN features. It can be observed that better manifold structure can be preserved after using our STMN method.
\section{Conclusions}\label{Sec.5}
We have proposed a spatio-temporal convolutional manifold network (STMN) to incorporate the manifold structure as a new constraint when extracting features using deep learning based approaches. Experimental results on two benchmark datasets demonstrated that our STMN method achieves competitive results for human action results. In future work, we will investigate how to combine our STMN method with the existing deep learning approaches such as two-stream CNN~\cite{Simoyan2014NIPS} and TSN~\cite{Wang2016ECCV} to further improve the recognition performance.

\bibliography{main}
\end{document}